\documentclass{article}

\PassOptionsToPackage{numbers, compress}{natbib}
\usepackage[preprint]{neurips_2026}

\usepackage[utf8]{inputenc} % allow utf-8 input
\usepackage[T1]{fontenc}    % use 8-bit T1 fonts
\usepackage{hyperref}       % hyperlinks
\usepackage{url}            % simple URL typesetting
\usepackage{booktabs}       % professional-quality tables (三线表)
\usepackage{amsfonts}       % blackboard math symbols
\usepackage{nicefrac}       % compact symbols for 1/2, etc.
\usepackage{microtype}      % microtypography
\usepackage{xcolor}         % colors
\usepackage{graphicx}       % images
\usepackage{multirow}       % 纵向合并单元格
\usepackage{amsmath}        % 数学公式

\title{DORA: Dynamic Online Reinforcement Agent for Token Merging in Vision Transformers}

\author{%
  Kaixuan He \\
  University of Science and Technology of China, Hefei, China \\
  Institute of Artificial Intelligence, Hefei Comprehensive National Science Center \\
  \texttt{hekaix@mail.ustc.edu.cn} \\
  \And
  Song Chen \\
  University of Science and Technology of China, Hefei, China \\
  \texttt{songch@ustc.edu.cn} \\  
  \And
  Yi Kang\thanks{Corresponding author.} \\
  University of Science and Technology of China, Hefei, China \\
  \texttt{ykang@ustc.edu.cn} \\
}

\begin{document}

\maketitle

\begin{abstract}
Vision Transformers (ViTs) incur significant computational overhead due to the quadratic complexity of self-attention relative to the token sequence length. While existing token reduction methods mitigate this issue, they predominantly rely on fixed heuristic metrics, predefined ratios, or static offline masks, which lack the adaptability to capture input-dependent redundancy during inference. In this paper, we propose \textbf{DORA} (\textbf{D}ynamic \textbf{O}nline \textbf{R}einforcement \textbf{A}gent), the first reinforcement learning (RL)-driven online inference framework for dynamic token merging in ViTs. We formulate the merging process as a sequential Markov Decision Process (MDP), where a lightweight RL agent determines the merging strategy for each Transformer block based on the current feature state and layer-specific context. To balance computational efficiency and feature fidelity, the agent is optimized via a dense reward function incorporating a non-linear distillation-based penalty. We implement an asymmetric Actor-Critic architecture that utilizes a high-capacity Critic for stable offline training while retaining a minimal Actor head for low-computation online inference. Evaluations across multiple ViT scales (Tiny to Large) demonstrate that DORA improves the accuracy-efficiency Pareto front compared to current baselines. Under strict negligible accuracy-drop constraints ($\le 0.05\%$), DORA achieves up to a 12.66\% token merging rate, and delivers up to a 569.7\% relative improvement over the most efficient baseline. On ImageNet-1K, under aligned accuracy constraints, DORA achieves up to a \textbf{76\% relative improvement} in computational savings compared to state-of-the-art methods. Furthermore, on out-of-distribution (OOD) benchmarks such as ImageNet-A and ImageNet-C, DORA attains a relative efficiency advantage of over \textbf{430\%}.
\end{abstract}

\section{Introduction}

Vision Transformers (ViTs)\cite{dosovitskiy2020image}\cite{touvron2021training}\cite{liu2021swin} are widely used in computer vision tasks, but their self-attention mechanism scales quadratically with the number of visual tokens, resulting in high computational complexity. Processing high-resolution inputs and dense token sequences presents challenges for deployment on resource-constrained devices and in real-time applications.

Token pruning and merging methods aim to reduce the number of tokens during the inference process. Existing approaches typically apply \textit{static budgeting} or \textit{fixed heuristics}. For example, methods using cosine similarity matching \cite{bolya2022tome}\cite{yoo2024adaptiv} or attention-based scoring \cite{rao2021dynamicvit}\cite{liang2022evit} execute token reduction under predefined computational budgets, such as fixed keep-ratios or constant layer-wise targets. More recently, offline reinforcement learning strategies \cite{vpruner} have emerged that learn pruning masks during the training phase and subsequently apply these static masks during inference; however, this approach lacks the ability to adapt dynamically to the varying characteristics of input data. Applying a uniform reduction intensity assumes constant redundancy across instances. When evaluated on out-of-distribution (OOD) shifts or image corruptions, these static methods exhibit performance drops, as the fixed reduction budget forces the discard of semantic features regardless of the input noise level.

In this paper, we propose \textbf{DORA} (\textbf{D}ynamic \textbf{O}nline \textbf{R}einforcement \textbf{A}gent) for dynamic token merging. DORA treats token reduction as an online sequential decision-making process, where the RL agent observes the state of the token sequence and dynamically generates a binary mask to execute token merging actions. The policy in DORA agent is optimized using a reward function that incorporates a non-linear fidelity penalty based on knowledge distillation from an unpruned teacher model, constraining semantic divergence while maximizing token reduction.

We evaluate DORA on standard benchmarks as well as OOD datasets. On ImageNet-1K, DORA achieves a 76\% relative improvement in computational savings and a significantly higher token merging rate compared to SOTA baselines like ToMe \cite{bolya2022tome} under aligned accuracy constraints. Furthermore, on out-of-distribution (OOD) benchmarks such as ImageNet-A and ImageNet-C, DORA expands the relative efficiency advantage to over \textbf{430\%} compared to heuristic methods, which fail to maintain effective reduction under environmental noise. 

Our primary contributions are summarized as follows:
\begin{itemize}
    \item We propose DORA, the first framework of the \textbf{online execution of a reinforcement learning agent} that can perform dynamic instance-by-instance token merging for vision transformers like ViT without predefined masks or thresholds.
    \item We model token reduction as a sequential decision-making process, and formulate the reduction process as a sequential Markov Decision Process where a lightweight RL agent determines the merging strategy for each Transformer block based on the current feature state.
    \item We evaluate token merging under more complex scenarios including not only standard image datasets but also those involving in distribution shifts and image corruptions.
\end{itemize}
The evaluation shows that DORA makes significant improvements over SOTA on token reduction rate, has higher robustness while incurring a negligible overhead.

\section{Related Work}

\subsection{Token Reduction in Vision Transformers}
The global self-attention mechanism in Vision Transformers scales quadratically with the number of input tokens. Token reduction techniques address this computational bottleneck through either pruning or merging. Pruning methods identify and discard uninformative patches \cite{rao2021dynamicvit, yin2022avit}. Merging approaches combine redundant tokens into unified representations to preserve the global information layout \cite{bolya2022tome, liang2022evit}. Both paradigms reduce theoretical FLOPs. Their reliance on static assumptions, such as fixed targets or local heuristics, restricts adaptability to varied input distributions.

\subsection{Heuristic and Budget-Constrained Compression}
Existing token reduction methods mostly rely on either fixed heuristics or predefined computational budgets, which significantly constrains their dynamic capacity.  

\paragraph{Heuristic Merging.} Methods such as ToMe \cite{bolya2022tome} and AdapTiV \cite{yoo2024adaptiv} based on cosine similarity to fuse tokens, where AdapTiV \cite{yoo2024adaptiv} utilizes local matching and a simplified sign similarity metric, lowering the computational overhead of the matching process itself. This similarity metric evaluates local feature distances. It does not account for the layer-specific semantic evolution of the network, which can result in sub-optimal token fusion in deeper blocks containing highly abstracted representations. PiToMe \cite{tran2024pitome} introduces an "energy score" to prioritize the preservation of informative tokens prior to merging, thereby offering enhanced protection for critical features in subsequent layers. DTEM \cite{lee2024dtem} learns a decoupled token embedding specifically for merging decisions, rather than relying directly on intermediate ViT features, which allows the merging strategy to be individually tailored and optimized.

\paragraph{Budget-Constrained Dynamic Pruning.} Dynamic scoring mechanisms incorporate auxiliary MLP predictors \cite{rao2021dynamicvit}, attention-guided sorting \cite{liang2022evit}, or zero-shot adaptive thresholding \cite{zerotprune}  to evaluate tokens. While these generated scores are input-dependent, the subsequent reduction process inherently relies on a \textit{predefined computational budget} or keep ratio ($\rho$) applied uniformly across all inputs. A constant reduction rate assumes identical informational density across different images. Under out-of-distribution (OOD) shifts or image corruptions, enforcing a strict keep ratio causes the inadvertent removal of structural semantic features alongside noise, leading to measured accuracy degradation.

\subsection{Reinforcement Learning for Network Compression}
Recently, V-Pruner \cite{vpruner} introduced RL to ViT token reduction. However, this approach is fundamentally \textit{offline}. It utilizes RL merely as an optimization tool during the training phase---based on a small batch of data---to search for a fixed token mask. Once deployed for inference, this static mask is rigidly applied, rendering the reduction strategy entirely input-independent. 

Unlike this offline method, our proposed \textbf{DORA} introduces an \textit{RL-driven online} merging paradigm. DORA eliminates the reliance on rigid \textit{predefined keep ratios} for token selection. During online inference, the RL agent observes the state of the token sequence at each block to generate a binary mask for the specific input. This input-dependent masking provides an adaptive mechanism for balancing computational efficiency with feature integrity.

\section{Methodology}

\subsection{Problem Formulation and Motivation}

Consider a Vision Transformer (ViT) consisting of $L$ sequential Transformer blocks. The input image is initialized as a sequence of $N$ visual tokens $\mathcal{X}_0=\{x_{0,1}, x_{0,2}, \dots, x_{0,N}\}$, where $x_{0,i}\in\mathbb{R}^d$. The self-attention mechanism processes these tokens with a computational complexity of $\mathcal{O}(N^2\cdot d)$, which becomes the primary bottleneck for real-time inference. 

Traditional token reduction methods typically simplify this as a static selection problem, where a fixed number of tokens are removed based on a predefined budget or heuristic metrics. Formally, they define a mapping $f: \mathcal{X}_l \rightarrow \mathcal{X}_{l+1}$ such that $|\mathcal{X}_{l+1}| = \lfloor \rho \cdot |\mathcal{X}_l| \rfloor$ or $|\mathcal{X}_{l+1}| = |\mathcal{X}_l| - r$, where $\rho$ and $r$ are constant hyper-parameters. However, such static paradigms ignore the fact that the optimal token density is highly dependent on the complexity of the input instance and the semantic stage of the network.

To overcome these limitations, we formulate the dynamic token merging process as a \textbf{Sequential Markov Decision Process (MDP)}, defined by the tuple $(\mathcal{S}, \mathcal{A}, \mathcal{P}, \mathcal{R}, \gamma)$. In our framework, DORA functions as an RL agent interacting with the ViT environment across $L$ decision steps:

\begin{itemize}
    \item \textbf{State Space $\mathcal{S}$}: At each block $l$, the state $s_l \in \mathcal{S}$ encapsulates the global semantic information of the current token sequence $\mathcal{X}_l$, augmented with the layer-specific context to capture the depth-wise evolution of features.
    \item \textbf{Action Space $\mathcal{A}$}: The action $a_l \in \mathcal{A}$ is a binary decision vector $M \in \{0, 1\}^N$, where each element indicates whether a token should be merged or preserved. Unlike static methods, the cardinality of the merged set is determined dynamically by the agent.
    \item \textbf{Transition Probability $\mathcal{P}$}: $\mathcal{P}(s_{l+1} | s_l, a_l)$ represents the environment's transition to the next block after performing the merging operation. This transition is deterministic in our setting, governed by the Transformer's forward pass.
    \item \textbf{Reward Function $\mathcal{R}$}: The reward $R_l(s_l, a_l)$ is designed to encourage maximal token reduction while minimizing the information loss relative to the original unpruned model's feature representations.
\end{itemize}

The objective of DORA is to learn an optimal policy $\pi(a_l | s_l)$ that maximises the expected cumulative reward $G = \sum_{l=1}^{L} \gamma^{l-1} R_l$. By modeling pruning as a sequential decision-making task, DORA can identify the "critical layers" for information aggregation and adapt its reduction intensity to the unique characteristics of each input instance.

\begin{figure}[htbp]
    \centering
    \includegraphics[width=1.0\textwidth]{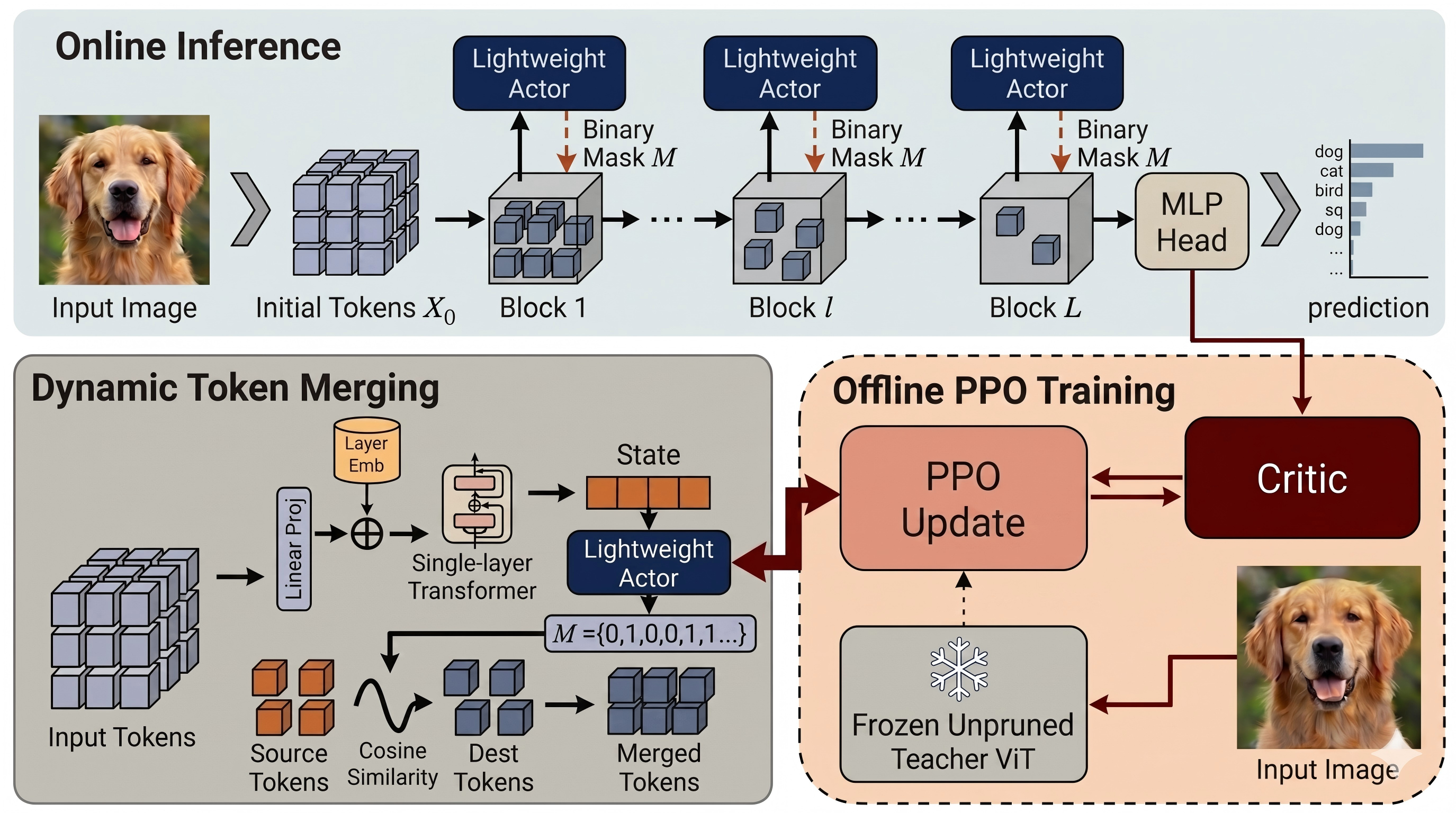} 
    \caption{\textbf{Overall architecture of the proposed DORA framework.} The framework employs a decoupled design, consisting of an offline training phase and an online inference phase. During the online phase, a pre-trained lightweight Actor network dynamically generates token merging masks, enabling input-adaptive acceleration for Transformers. In the offline phase, a high-capacity Critic network assists in training and optimizing the Actor's policy via the PPO algorithm \cite{schulman2017proximal} and a reward based on knowledge distillation. The Critic is discarded after training.}
    \label{fig:architecture}
\end{figure}

\subsection{State and Action Representation}

To enable input-dependent and layer-adaptive decision-making, we design the state and action representations to balance computational efficiency with representational richness.

\subsubsection{Depth-Aware State Representation}
In our framework, the RL agent does not directly operate on the raw high-dimensional tokens. Instead, it extracts a \textit{context-aware state representation} $S_l$ to inform its merging decisions. Given the input token sequence $\mathcal{X}_l \in \mathbb{R}^{N \times d}$, we first apply a linear projection $\text{Proj}(\mathcal{X}_l)$ to reduce the feature dimension, filtering redundant information. 
A learnable layer embedding $\text{Emb}(l)$ is added to the projected sequence to encode the block index:
\begin{equation}
    \text{Emb}(l) = \mathbf{E}_{[l, :]}
\end{equation}
where $\mathbf{E} \in \mathbb{R}^{L \times d'}$ is a learnable embedding matrix encompassing all $L$ Transformer blocks, and $\mathbf{E}_{[l, :]}$ extracts the specific $d'$-dimensional positional vector for the $l$-th block. 

The final state $S_l$ is obtained by passing this augmented sequence through a lightweight Transformer encoder:
\begin{equation}
    S_l = \text{TransformerEncoder}\left( \text{Proj}(\mathcal{X}_l) + \text{Emb}(l) \right)
\end{equation}
where $S_l \in \mathbb{R}^{N \times d'}$ represents the extracted state at block $l$. This design ensures that $S_l$ not only captures the unique characteristics of the current instance through $\text{Proj}(\mathcal{X}_l)$ but also encodes the semantic stage of the network via $\text{Emb}(l)$. Consequently, the Actor head can generate precise, input-dependent merging actions based on this refined state representation.

This depth-aware representation enables layer-adaptive token reduction. By explicitly conditioning on the layer index $l$, the policy adapts to the non-uniform distribution of token redundancy across the network. Experimental results show a depth-dependent reduction pattern. While the policy retains a larger number of tokens in the earlier layers, the merging rate increases in the deeper blocks, a shift that correlates with the aggregation of global features. For example, ViT-Tiny exhibits peak merging activity at blocks 6, 8, and 9. The layer embedding provides the necessary positional signal for such layer-adaptive temporal specialization.

\subsubsection{Dynamic Token Merging Action}
For each token $x_{l,i} \in \mathcal{X}_l$, the Actor network within DORA outputs a binary decision $m_i \in \{0, 1\}$, forming an input-dependent mask $M = \{m_1, m_2, \dots, m_N\}$. Unlike prior works that rely on a fixed budget $\rho$, the density of $M$ in DORA is entirely determined by the policy $\pi(a_l | s_l)$ in response to the specific input image.

The merging process is then executed via an \textbf{asymmetric many-to-one aggregation} mechanism. Specifically, we partition the current token set into a source set $\mathcal{I}_{src} = \{i \mid m_i = 1\}$ (tokens to be merged) and a destination set $\mathcal{I}_{dst} = \{j \mid m_j = 0\}$ (tokens to be preserved). For each source token $i \in \mathcal{I}_{src}$, the agent identifies its most semantically similar counterpart in the destination set based on cosine similarity:
\begin{equation}
    j^* = \arg\max_{j \in \mathcal{I}_{dst}} \frac{x_{l,i} \cdot x_{l,j}}{\|x_{l,i}\| \|x_{l,j}\|}
\end{equation}
The features of the source token $x_{l,i}$ are then aggregated into $x_{l,j^*}$ via a weighted average. Unlike strict bipartite matching which enforces one-to-one constraints, our asymmetric formulation allows multiple redundant tokens (e.g., from extensive background regions) to be concurrently fused into a single representative destination token. This asymmetric mapping aggregates all source features into the destination set, avoiding explicit token deletion. This mechanism enables variable reduction rates across different spatial regions.

\subsection{Dense Reward Shaping for Dynamic Pruning}

Formulating the reward function is a critical factor in RL-based token reduction. Sparse reward signals frequently trigger policy collapse, resulting in boundary behaviors such as complete sequence preservation or maximum token removal. To stabilize policy optimization, DORA implements a step-wise reward mechanism. This reward is regularized by a Knowledge Distillation (KD) \cite{hinton2015distilling} constraint, utilizing feature representations from a frozen, unpruned teacher model.

At each Transformer block $l$, the agent receives a step-wise reward $R_{step, l}$. The reward is defined by the number of successfully merged tokens, encouraging the agent to sparsify the sequence. To prevent excessive feature divergence, we introduce a non-linear fidelity penalty:
\begin{equation}
\label{eq:reward}
    R_{step, l} = \alpha \cdot n_{merge, l} - \beta \cdot \max(0, \Delta \mathcal{L}_{KD, l})^2
\end{equation}
where $n_{merge, l}$ denotes the number of tokens merged at block $l$, and $\Delta \mathcal{L}_{KD, l}$ represents the incremental KD loss (feature divergence) induced by the current merging action compared to the teacher model. $\alpha$ and $\beta$ are scaling factors.

The squared penalty term, $\max(0, \Delta \mathcal{L}_{KD, l})^2$, functions as a non-linear constraint on the action space. Minor feature deviations yield minimal scalar penalties, permitting higher reduction rates in regions with low representation variance. Conversely, significant deviations from the teacher model incur a quadratic penalty. This mechanism regularizes the policy, preventing actions that heavily alter the original feature structure.

\subsection{Asymmetric Actor-Critic Architecture}

DORA employs a decoupled, \textbf{Asymmetric Actor-Critic} \cite{jiang2026asymmetric} architecture.

The framework implements an asymmetric parameter allocation between the Actor and Critic networks. During offline training, the Critic utilizes expanded projection dimensions to compute advantage estimations. It is discarded post-training and excluded from inference.

The Actor network is the sole RL component retained for online deployment. Independent of the base ViT architecture size, the Actor projects the input tokens into a 64-dimensional latent space. This state is subsequently processed by a single-layer Transformer encoder. To address the structural discrepancy between the two networks, the Actor and Critic are optimized using decoupled learning rates. This configuration restricts runtime computational overhead to the Actor head while leveraging the larger parameter space of the Critic to stabilize policy gradient updates.

\section{Experimental Results}

\subsection{Experimental Setup}

\textbf{Datasets and Evaluation Metrics.} We evaluate the performance of DORA on the standard ImageNet-1K classification benchmark \cite{deng2009imagenet}. To further assess the dynamic adaptation and robustness of our reinforcement learning agent, we conduct comprehensive testing on multiple out-of-distribution (OOD) benchmarks. These include \textbf{ImageNet-A} \cite{hendrycks2021natural}, which contains natural adversarial examples; \textbf{ImageNet-R} \cite{hendrycks2021many}, which features various artistic renditions and stylized images; and three macro-categories (\textbf{Weather}, \textbf{Blur}, and \textbf{Digital}) of \textbf{ImageNet-C} \cite{hendrycks2019benchmarking}, which encompass a wide range of algorithmic and environmental image corruptions. The primary evaluation metrics are Top-1 accuracy, the absolute FLOPs of the pruned models, and the average token reduction rate per block.

\textbf{Model Architectures and Pre-trained Weights.} We evaluate two architectures: the \textbf{ViT} \cite{dosovitskiy2020image} (Tiny, Small, Base, Large) using \textbf{AugReg} pre-trained weights \cite{steiner2021how}, and the \textbf{DeiT} \cite{touvron2021training} (Tiny, Small, Base) using \textbf{hard-distillation variants}.

\textbf{Baselines.} We benchmark DORA against established token reduction baselines, categorized into heuristic matching approaches (ToMe \cite{bolya2022tome}), learned dynamic pruning methods (ZeroTPrune \cite{zerotprune}), and offline RL strategies (V-Pruner \cite{vpruner}). All baselines are re-implemented under a unified setup to ensure fair comparison.

\subsection{Main Results on ImageNet-1K}
\label{sec:main_results}

On the ImageNet-1K validation set, we adjust baseline hyperparameters to align with DORA's Top-1 accuracy ($\pm 0.05\%$), isolating token compression as the primary metric under equivalent performance constraints.

\textbf{Average Token Reduction Rate.} Table \ref{tab:main_results} reports the absolute GFLOPS and token compression metrics. The \textit{Tokens Reduction (\%)} column indicates the average block-wise token reduction rate, defined as the proportion of reduced tokens across all Transformer layers relative to the unpruned sequence length. Under aligned accuracy constraints, DORA achieves the highest token reduction rates across all evaluated architectures.

\begin{figure}[htbp]
    \centering
    \includegraphics[width=1.0\textwidth]{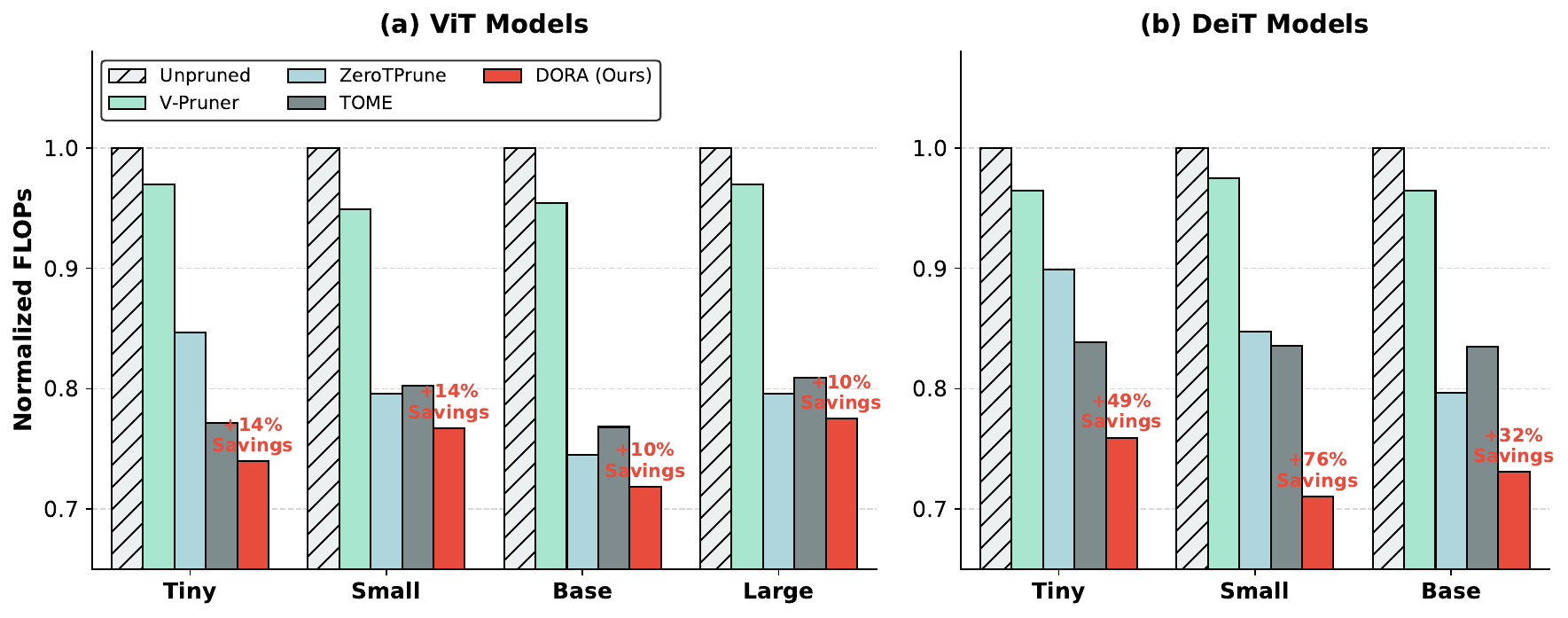}
    \caption{\textbf{Computational savings comparison of token reduction methods under strictly aligned Top-1 accuracy constraints.} \textbf{(a)} Normalized FLOPs results on the ViT architectures. \textbf{(b)} Normalized FLOPs results on the DeiT architectures.}
    \label{fig:flops_normalized}
\end{figure}

\textbf{Normalized FLOPs and Relative Savings.} Figure \ref{fig:flops_normalized} reports the normalized FLOPs of each method relative to the unpruned baselines. To quantify comparative efficiency, the figure annotates the \textbf{additional relative savings} (denoted as \textit{+XX\% Savings}) of DORA over the most efficient baseline at each model scale. These savings are derived from net FLOPs, incorporating the computational overhead of both the DORA method and the baseline matching algorithms. For the ViT architectures, DORA yields an additional \textbf{10\% to 14\%} relative reduction in computational cost. On the DeiT variants, these relative savings expand to a range of \textbf{32\% to 76\%}.

\begin{table}[htbp]
    \centering
    \caption{\textbf{Quantitative comparison of token reduction methods on ImageNet-1K.} We align the Top-1 accuracy of all methods to ensure a fair comparison of computational efficiency. The highest token reduction rate and lowest GFLOPS under identical accuracy constraints are highlighted in \textbf{bold}.}
    \label{tab:main_results}
    
    % 可选：将表格内的字体稍微调小一号（如 \small 或 \footnotesize）
    \small 
    
    % 使用 resizebox 将表格强制缩放至文本宽度 (\textwidth)
    \resizebox{\textwidth}{!}{%
    \begin{tabular}{llcccccc}
        \toprule
        \multirow{2}{*}{\textbf{Scale}} & \multirow{2}{*}{\textbf{Method}} & \multicolumn{3}{c}{\textbf{ViT}} & \multicolumn{3}{c}{\textbf{DeiT}} \\
        \cmidrule(lr){3-5} \cmidrule(lr){6-8}
        & & \textbf{Acc (\%)} & \textbf{GFLOPS} & \textbf{Tokens Red. (\%)} & \textbf{Acc (\%)} & \textbf{GFLOPS} & \textbf{Tokens Red. (\%)} \\
        \midrule
        
        % ================= Tiny =================
        \multirow{5}{*}{\textbf{Tiny}} 
        & Base (Unpruned) & 75.5 & 1.25 & 0.0 & 74.5 & 1.25 & 0.0 \\
        & V-Pruner \cite{vpruner} & 74.9 & 1.22 & 3.1 & 74.2 & 1.21 & 3.6 \\
        & ZeroTPrune \cite{zerotprune} & 74.9 & 1.06 & 15.4 & 74.2 & 1.13 & 10.1 \\
        & ToMe \cite{bolya2022tome} & 74.9 & 0.97 & 23.2 & 74.2 & 1.05 & 16.6 \\
        & \textbf{DORA (Ours)} & \textbf{74.9} & \textbf{0.93} & \textbf{28.5} & \textbf{74.2} & \textbf{0.95} & \textbf{26.6} \\
        \midrule
        
        % ================= Small =================
        \multirow{5}{*}{\textbf{Small}} 
        & Base (Unpruned) & 81.4 & 4.60 & 0.0 & 81.2 & 4.60 & 0.0 \\
        & V-Pruner \cite{vpruner} & 81.0 & 4.37 & 5.1 & 81.0 & 4.49 & 2.5 \\
        & ZeroTPrune \cite{zerotprune} & 81.0 & 3.66 & 20.4 & 81.0 & 3.90 & 15.3 \\
        & ToMe \cite{bolya2022tome} & 81.0 & 3.69 & 19.9 & 81.0 & 3.85 & 16.6 \\
        & \textbf{DORA (Ours)} & \textbf{81.0} & \textbf{3.53} & \textbf{24.1} & \textbf{81.0} & \textbf{3.27} & \textbf{29.7} \\
        \midrule

        % ================= Base =================
        \multirow{5}{*}{\textbf{Base}} 
        & Base (Unpruned) & 84.6 & 17.58 & 0.0 & 83.3 & 17.58 & 0.0 \\
        & V-Pruner \cite{vpruner} & 84.0 & 16.78 & 4.6 & 83.1 & 16.96 & 3.6 \\
        & ZeroTPrune \cite{zerotprune} & 84.0 & 13.10 & 25.5 & 83.1 & 14.01 & 20.3 \\
        & ToMe \cite{bolya2022tome} & 84.0 & 13.51 & 23.2 & 83.1 & 14.67 & 16.6 \\
        & \textbf{DORA (Ours)} & \textbf{84.0} & \textbf{12.64} & \textbf{28.4} & \textbf{83.1} & \textbf{12.85} & \textbf{27.2} \\
        \midrule
        
        % ================= Large =================
        \multirow{5}{*}{\textbf{Large}} 
        & Base (Unpruned) & 85.8 & 61.61 & 0.0 & - & - & - \\
        & V-Pruner \cite{vpruner} & 85.6 & 59.73 & 3.1 & - & - & - \\
        & ZeroTPrune \cite{zerotprune} & 85.6 & 49.02 & 20.4 & - & - & - \\
        & ToMe \cite{bolya2022tome} & 85.6 & 49.84 & 19.1 & - & - & - \\
        & \textbf{DORA (Ours)} & \textbf{85.6} & \textbf{47.74} & \textbf{22.7} & - & - & - \\
        \bottomrule
    \end{tabular}%
    } % 注意这里的 } 闭合了 resizebox
\end{table}

\begin{table}[htbp]
    \centering
    \caption{\textbf{Token reduction performance under negligible accuracy-drop constraints.} We report the average token reduction rate (\%) achieved by each method while maintaining a strict accuracy drop constraint ($\Delta \text{Top-1 Acc} \le 0.05\%$). The last column denotes the relative gain of DORA compared to the highest-performing baseline. The best results under this negligible-degradation setting are highlighted in \textbf{bold}.}
    \label{tab:lossless_results}
    \small
    % 使用 tabular* 和 \extracolsep{\fill} 来实现等比例撑满整个页面宽度
    \begin{tabular*}{\textwidth}{@{\extracolsep{\fill}}llcccc@{}}
        \toprule
        \multirow{2}{*}{\textbf{Architecture}} & \multirow{2}{*}{\textbf{Model Scale}} & \multicolumn{4}{c}{\textbf{Tokens Reduction (\%)}} \\
        \cmidrule(lr){3-6}
        & & \textbf{ZeroTPrune \cite{zerotprune}} & \textbf{ToMe \cite{bolya2022tome}} & \textbf{DORA (Ours)} & \textbf{Rel. Gain} \\
        \midrule
        \multirow{4}{*}{\textbf{ViT}} 
        & Tiny  & 1.52\% & 0.00\% & \textbf{10.18\%} & \textbf{+569.7\%} \\
        & Small & 5.33\% & 3.32\% & \textbf{10.26\%} & \textbf{+92.5\%} \\
        & Base  & 5.33\% & 6.63\% & \textbf{9.09\%}  & \textbf{+37.1\%} \\
        & Large & 4.57\% & 6.38\% & \textbf{7.94\%}  & \textbf{+24.5\%} \\
        \midrule
        \multirow{3}{*}{\textbf{DeiT}} 
        & Tiny  & 3.05\% & 6.63\% & \textbf{11.14\%} & \textbf{+68.0\%} \\
        & Small & 3.05\% & 3.32\% & \textbf{12.66\%} & \textbf{+281.3\%} \\
        & Base  & 3.81\% & 6.63\% & \textbf{12.13\%} & \textbf{+83.0\%} \\
        \bottomrule
    \end{tabular*}
\end{table}

\paragraph{Performance under Negligible Accuracy-Drop Constraints.} Under a strict maximum accuracy degradation constraint of 0.05\% (Table \ref{tab:lossless_results}), heuristic and fixed-budget methods yield minimal token reduction. Their static reduction mechanisms fuse target-class feature vectors even at low reduction intensities, causing immediate accuracy drops. DORA maintains baseline accuracy while achieving a maximum token reduction rate of \textbf{12.66\%}. On the ViT-Tiny architecture, it delivers a maximum relative improvement of \textbf{+569.7\%} over the strongest competing baseline.This indicates that the input-dependent policy isolates and merges regions of homogeneous backgrounds without altering the structural features required for high-fidelity inference.

\subsection{Robustness to Out-of-Distribution Data}
\label{sec:ood_robustness}

The robustness of the RL-driven framework is evaluated on multiple OOD datasets: \textbf{ImageNet-A} \cite{hendrycks2021natural}, \textbf{ImageNet-R} \cite{hendrycks2021many}, and \textbf{ImageNet-C} \cite{hendrycks2019benchmarking}. Evaluations are conducted under two comparative constraints: aligned token reduction rates and aligned Top-1 accuracy.

\begin{table}[htbp]
    \centering
    \caption{\textbf{Comprehensive robustness evaluation on Out-of-Distribution (OOD) datasets.} We evaluate the methods across natural adversarial examples (ImageNet-A), renditions (ImageNet-R), and three macro-categories of corruptions (ImageNet-C). Under equivalent token reduction rates, DORA consistently maintains higher accuracy. When constrained to match DORA's accuracy, the heuristic baseline (ToMe\cite{bolya2022tome}) suffers severe degradation in its token reduction capabilities.}
    \label{tab:ood_results_comprehensive}
    \small
    % 使用 tabular* 撑满页面宽度，使得排版更大气
    \begin{tabular*}{\textwidth}{@{\extracolsep{\fill}}llcccc@{}}
        \toprule
        \multirow{2}{*}{\textbf{OOD Benchmark}} & \multirow{2}{*}{\textbf{Method}} & \multicolumn{2}{c}{\textbf{Similar Reduction Rate}} & \multicolumn{2}{c}{\textbf{Aligned Accuracy}} \\
        \cmidrule(lr){3-4} \cmidrule(lr){5-6}
        & & \textbf{Tokens Red. (\%)} & \textbf{Acc (\%)} & \textbf{Acc (\%)} & \textbf{Tokens Red. (\%)} \\
        \midrule
        
        % ================= 独立 OOD 数据集 =================
        \multicolumn{6}{l}{\textit{Adversarial \& Stylized Datasets}} \\
        \addlinespace[0.3em]
        \multirow{2}{*}{\textbf{ImageNet-A}} 
        & ToMe \cite{bolya2022tome} & 33.16 & 8.44 & 8.99 & 0.00 \\
        & \textbf{DORA (Ours)} & \textbf{32.86} & \textbf{9.51} ($\uparrow$) & \textbf{9.51} & \textbf{32.86} ($\uparrow$) \\
        \addlinespace[0.2em]
        \multirow{2}{*}{\textbf{ImageNet-R}} 
        & ToMe \cite{bolya2022tome} & 23.21 & 31.66 & 31.74 & 0.00 \\
        & \textbf{DORA (Ours)} & \textbf{25.25} & \textbf{32.03} ($\uparrow$) & \textbf{32.03} & \textbf{25.25} ($\uparrow$) \\
        \midrule

        % ================= ImageNet-C 宏观分类 =================
        \multicolumn{6}{l}{\textit{ImageNet-C Corruptions (Macro-Averages)}} \\
        \addlinespace[0.3em]
        \multirow{2}{*}{\textbf{IN-C (Weather)}} 
        & ToMe \cite{bolya2022tome} & 37.81 & 40.65 & 42.63 & 7.30 \\
        & \textbf{DORA (Ours)} & \textbf{38.98} & \textbf{43.33} ($\uparrow$) & \textbf{43.33} & \textbf{38.98} ($\uparrow$) \\
        \addlinespace[0.2em]
        \multirow{2}{*}{\textbf{IN-C (Blur)}} 
        & ToMe \cite{bolya2022tome} & 39.80 & 33.91 & 36.19 & 18.57 \\
        & \textbf{DORA (Ours)} & \textbf{40.68} & \textbf{36.22} ($\uparrow$) & \textbf{36.22} & \textbf{40.68} ($\uparrow$) \\
        \addlinespace[0.2em]
        \multirow{2}{*}{\textbf{IN-C (Digital)}} 
        & ToMe \cite{bolya2022tome} & 39.80 & 46.49 & 48.59 & 20.73 \\
        & \textbf{DORA (Ours)} & \textbf{40.73} & \textbf{48.48} ($\uparrow$) & \textbf{48.48} & \textbf{40.73} ($\uparrow$) \\
        \bottomrule
    \end{tabular*}
\end{table}

\textbf{Performance under Similar Reduction Rates.} Table \ref{tab:ood_results_comprehensive} reports the evaluation results. Under constrained token reduction rates, DORA maintains higher Top-1 accuracy compared to the heuristic baseline across the tested corruption types. This performance indicates that the learned policy filters tokens based on input-dependent structural features rather than relying on standard dataset background patterns. In the presence of input corruption, the agent isolates noise-induced tokens for reduction while preserving the feature vectors of the target class.

\textbf{Performance under Aligned Accuracy.} The disparity in token selection efficiency becomes apparent when baseline methods are constrained to match the Top-1 accuracy of DORA. To maintain equivalent accuracy on ImageNet-C, the token reduction rate of ToMe decreases to between 7.3\% and 20.73\%. In contrast, DORA sustains an approximately 38\% to 40\% token reduction rate. Specifically, on the ImageNet-C (Weather) subset, DORA's 38.98\% reduction rate compared to ToMe's 7.30\% translates to a \textbf{434\% relative improvement} in compression efficiency. Furthermore, on the ImageNet-A and ImageNet-R datasets, the baseline does not attain DORA's accuracy levels (9.51\% and 32.03\%, respectively) even when token reduction is completely disabled (0.0\% merged). Conversely, the RL policy achieves these accuracies while discarding 32.86\% and 25.25\% of the tokens, respectively. These results demonstrate that dynamic token merging functions as an effective noise-filtering mechanism for processing OOD data.

\subsection{Visualization and Behavioral Patterns}
\label{sec:analysis}

\begin{figure}[htbp]
    \centering
    \includegraphics[width=1.0\textwidth]{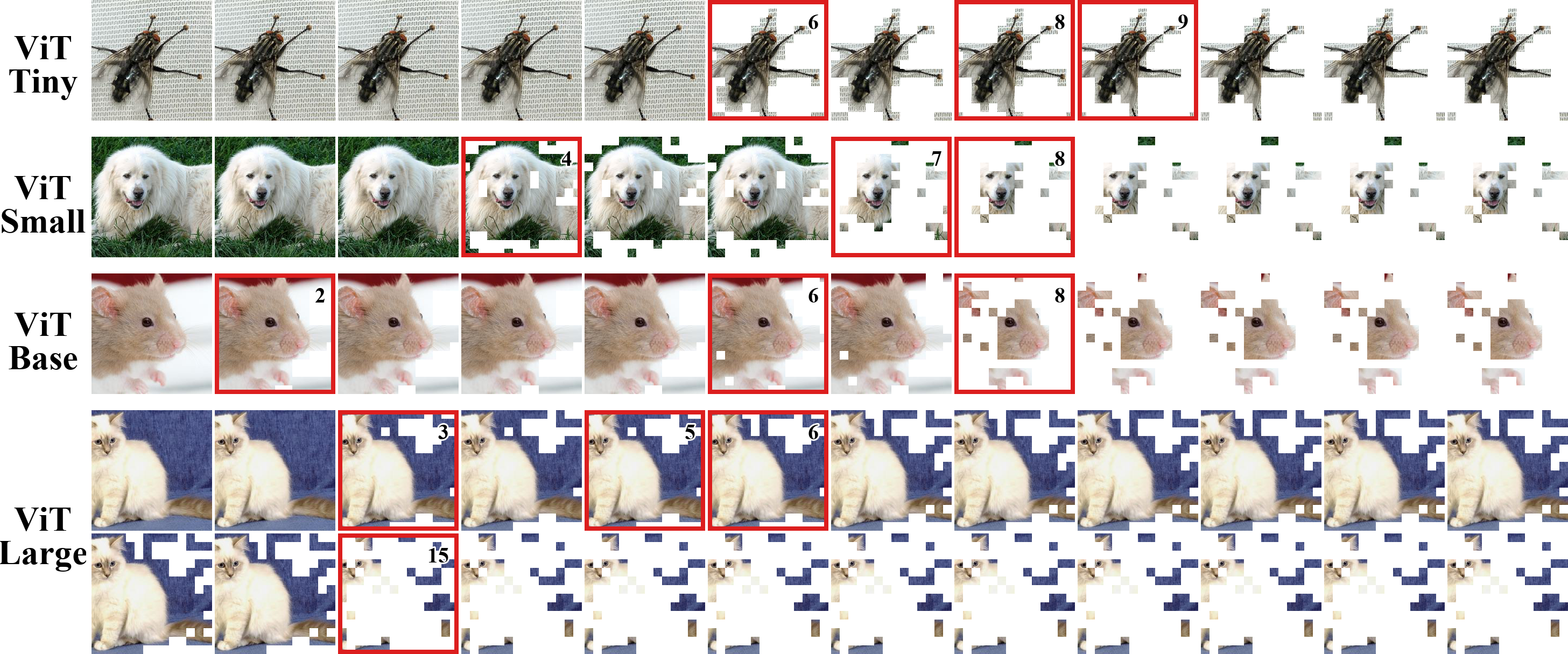}
    \caption{\textbf{Visualization of the dynamic merging process.} Red bounding boxes indicate the specific Transformer blocks where the RL agent executes token merging, with the corresponding block index annotated in the top-right corner. Unmarked images represent blocks where the agent opted for zero-merging actions (bypassing the reduction step).}
    \label{fig:visualization_all}
\end{figure}

Figure \ref{fig:visualization_all} visualizes the token merging process across multiple Transformer blocks for the ViT-Tiny, Small, Base, and Large architectures. The sequential sub-images record the spatial distribution of the retained tokens after each block. The visualizations indicate a \textbf{layer-adaptive} merging pattern. Token retention is maximized in the early blocks, corresponding to the processing of local geometric features. Higher merging rates occur in specific deeper blocks, aligning with the network's transition toward global semantic representations. For instance, the agent actively triggers merging at blocks 6, 8, and 9 for ViT-Tiny, and at blocks 3, 5, 6, and 15 for ViT-Large. This non-uniform distribution of merging actions supports the formulation of token merging as a sequential Markov Decision Process (MDP) rather than a constant-ratio schedule.

\subsection{Computational Overhead Analysis}
\label{sec:overhead}

The layer-wise sparsity mentioned in Section \ref{sec:analysis} facilitates inference acceleration. As evidenced by the unmarked blocks in Figure \ref{fig:visualization_all}, the policy frequently generates zero-merging actions. At these stages, the Actor network's forward pass can be completely bypassed to reduce computational overhead. Table \ref{tab:overhead} details the computational overhead of the DORA method relative to the unpruned backbones. DORA accounts for \textbf{less than 1\%} of the total FLOPS for the Small, Base, and Large models, and 2.59\% for the Tiny baseline. The FLOPs reported in Section \ref{sec:main_results} incorporate this internal decision-making cost.

\begin{table}[htbp]
    \centering
    \caption{\textbf{Inference overhead analysis of the DORA method.} We report the additional GFLOPS and the percentage relative to the original models.}
    \label{tab:overhead}
    \small
    \begin{tabular*}{\textwidth}{@{\extracolsep{\fill}}lccc@{}}
        \toprule
        \textbf{Model Scale} & \textbf{Original GFLOPS} & \textbf{Overhead (GFLOPS)} & \textbf{Overhead Ratio (\%)} \\
        \midrule
        Tiny  & 1.25  & 0.032 & 2.59\% \\
        Small & 4.60  & 0.038 & 0.83\% \\
        Base  & 17.58 & 0.050 & 0.28\% \\
        Large  & 61.61 & 0.097 & 0.16\% \\
        \bottomrule
    \end{tabular*}
\end{table}

\section{Conclusion and Future Work}
\label{sec:conclusion}

This paper presents \textbf{DORA}, an online reinforcement learning framework designed for input-dependent token merging in Vision Transformers. By modeling the merging process as a sequential decision-making task, DORA replaces static heuristics with a policy that adapts to the informational density of individual inputs. Experimental results on ImageNet-1K show that DORA achieves up to a \textbf{76\% relative improvement} in computational savings under aligned accuracy constraints. Furthermore, DORA maintains its merging rate under distribution shifts; on multiple out-of-distribution (OOD) benchmarks, the framework sustains high merging rates where heuristic baselines require significant reduction in pruning to preserve accuracy. These findings demonstrate that dynamic, learned token merging effectively balances computational efficiency with model robustness.

Future research will focus on extending the RL-driven merging paradigm to \textbf{Video Transformers} \cite{bertasius2021spacetime}\cite{liu2022video} to evaluate its performance on structured, long-range dependencies. Additionally, we plan to investigate \textbf{multi-agent reinforcement learning} configurations to jointly optimize pruning and merging operations across different network stages, aiming to further refine the efficiency-accuracy trade-offs in dynamic neural architectures.

\bibliographystyle{unsrtnat}
\bibliography{references}

\newpage

\end{document}